\documentclass{article}

\usepackage{times}
\usepackage{graphicx} 
\usepackage{subfigure} 

\usepackage{natbib}

\usepackage{algorithm}
\usepackage{algorithmic}

\usepackage{hyperref}

\usepackage{bm}
\usepackage{amssymb}

\usepackage[accepted]{icml2017} 

\icmltitlerunning{Actor-Critic Reinforcement Learning with Simultaneous Human Control and Feedback}

\title{Actor-Critic Reinforcement Learning with Simultaneous Human Control and Feedback}

\begin{document} 

\twocolumn[
\icmltitle{Actor-Critic Reinforcement Learning with \\ Simultaneous Human Control and Feedback}



\icmlsetsymbol{equal}{*}

\begin{icmlauthorlist}
\icmlauthor{Kory W. Mathewson}{ua}
\icmlauthor{Patrick M. Pilarski}{uam}
\end{icmlauthorlist}

\icmlaffiliation{ua}{University of Alberta, Dep. of Computing Science, Edmonton, Canada}
\icmlaffiliation{uam}{University of Alberta, Deps. of Medicine and Computing Science, Edmonton, Alberta, Canada}
\icmlcorrespondingauthor{Kory Mathewson}{korym@ualberta.ca}

\icmlkeywords{human-robot interaction, control and feedback, actor-critic, reinforcement learning, learning from demonstration, face-valuing, machine learning}

\vskip 0.3in
]



\printAffiliationsAndNotice{}  

\begin{abstract} 

This paper contributes a first study into how different human users deliver simultaneous control and feedback signals during human-robot interaction. As part of this work, we formalize and present a general interactive learning framework for online cooperation between humans and reinforcement learning agents. In many human-machine interaction settings, there is a growing gap between the degrees-of-freedom of complex semi-autonomous systems and the number of human control channels. Simple human control and feedback mechanisms are required to close this gap and allow for better collaboration between humans and machines on complex tasks. To better inform the design of concurrent control and feedback interfaces, we present experimental results from a human-robot collaborative domain wherein the human must simultaneously deliver both control and feedback signals to interactively train an actor-critic reinforcement learning robot. We compare three experimental conditions: 1) human delivered control signals, 2) reward-shaping feedback signals, and 3) simultaneous control and feedback. Our results suggest that subjects provide less feedback when simultaneously delivering feedback and control signals and that control signal quality is not significantly diminished. Our data suggest that subjects may also modify when and how they provide feedback. Through algorithmic development and tuning informed by this study, we expect semi-autonomous actions of robotic agents can be better shaped by human feedback, allowing for seamless collaboration and improved performance in difficult interactive domains.

\end{abstract} 

\begin{figure}[h!]
	\centering
    \includegraphics[width=3.3in]{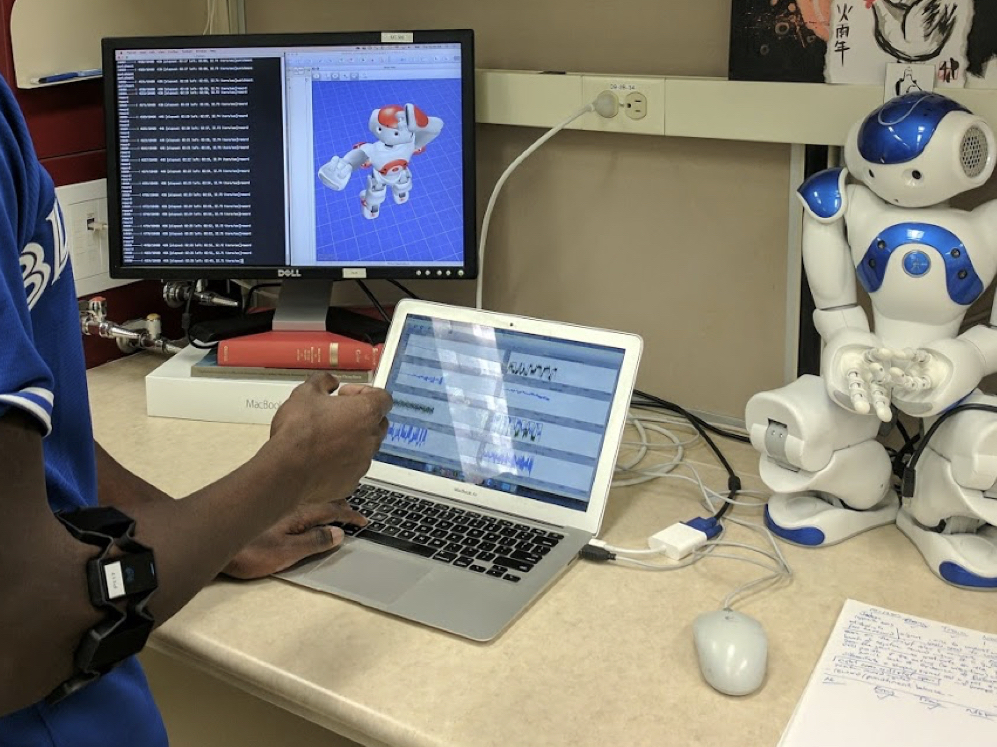}
  	\caption{Experimental configuration. One of the study participants with the Myo band on their right arm providing a control signal, while simultaneously providing feedback signals with their left hand. The Aldebaran Nao robot simulation is visible on the screen alongside experimental logging.}
    \label{fig:fig7}
\end{figure}

\section{Introduction}
\label{intro}

Interactive machine learning (IML) is the field of computing science exploring how intelligent agents solve tasks together. Complex tasks of the future will demand adaptive collaboration at the interface between human and machine intelligences over the course of sequential decision-making problems. These problem domains span many complex tasks: from search and recommendations systems to social media personalization and safety, from interactive dialogue systems to embodied agents, and from autonomous vehicles to prosthetic robots. 

In this work we focus on IML in the setting of human-machine interfaces (HMI) where the human and the machine interact closely and frequently with each other, e.g., a human interacting with a robotic artificial limb. Tightly coupled HMI of this kind have become increasingly complex in form and function, and, in critical domains such as prosthetics, control remains a serious challenge for users \cite{Castellini2014ProceedingsElectromyography}. Advances in engineering have resulted in an increasing number of controllable actuators and a resulting chasm between these actuators and the limited number of clean, reproducible control signals from a user. 

\begin{figure*}[t]
	\centering
    \includegraphics[width=0.9\textwidth]{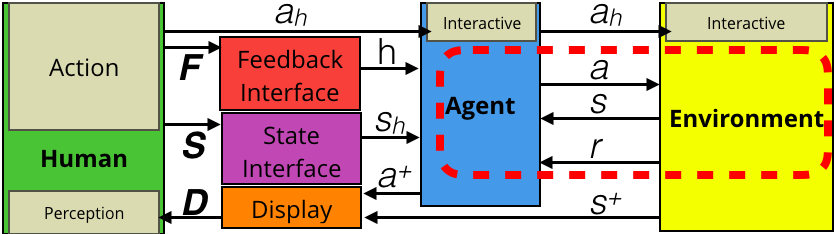}
  	\caption{General Interactive Learning Framework. This diagram shows the interactions between the prepared human and the agent (learning system) during task performance. The system is composed of three dense communication channels: $\bm{F}$: feedback signal from the human, processed by Feedback Interface, $\bm{S}$: state signal from the human processed by State Interface, $\bm{D}$: display signals from agent action information and environmental state information processed by a Display to the perception of the human; several signals from the human: h: real-valued human signals delivered to agent, $a_h$: human actions which may impact interactive components of the agent or environment directly, $s_h$: human state signals delivers to agent; two meta-information signals: $a^+$: information about action taken by learning agent, and $s^+$: information about the new state of the environment; and the basic RL diagram (denoted in the red dashed box): $a$: action performed by agent in environment, $s$: state of the environment, delivered to the agent, and $r$: reward delivered by the environment to the agent}
    \label{fig:fig8}
\end{figure*}


One flexible way to manage the growing disparity between available control options and accessible human control signals is to transfer partial autonomy to machine learning agents. IML research to date has investigated how autonomous agents can learn to solve problems effectively through interactions with humans \cite{argall2009survey, Knox2009InteractivelyFramework, fails2003interactive}. Seamless communication between autonomous agents and humans demands two-way communication and understanding. Past IML research has therefore explored techniques ranging from human demonstrations, to human-robotic control sharing and humans delivering feedback \cite{argall2009survey, Knox2015FramingPerformance, fails2003interactive}.

In this paper we  focus specifically on humans interacting with reinforcement learning (RL) systems  \cite{thomaz2005real, Knox2015FramingPerformance}. Of note, Actor-Critic Reinforcement Learning (ACRL) is a family of RL algorithms which have shown promise in past studies as a way for humans and autonomous agents to collaborate in tightly coupled HMI like prosthetic limbs and other robots \cite{Pilarski2013Real-timeJoints,Pilarski2011OnlineLearning}. 





While studies with IML have explored in detail the ways in which humans deliver feedback to a machine learner \cite{Thomaz2008TeachableLearners,Knox2015FramingPerformance}, it remains unclear to what degree humans are able to provide {\em both} feedback and primary control signals to a machine partner. It is also unclear what interactions exist between a human providing tightly coupled direct control signals and high frequency reward and punishment signals to a learning robotic system. Simultaneous control and feedback is a novel, unexplored domain with many unanswered questions in the space of human-RL interaction, specifically when multi-model signals must be interpreted by the learning system.

This paper therefore contributes the first study into \textit{simultaneous human control and feedback} during the training of an ACRL algorithm. To support this study, we present a General Interactive Learning Framework (Fig. \ref{fig:fig2}) within which these problems can be easily described and formalized. We then present results from a user study (N=13) on a well-defined testbed for experimentation in the space of IML for humans and RL algorithms with control and feedback. Finally, we conclude by discussing open questions and propose several future research topics in interactive RL.

\section{Background}

RL is a learning framework inspired by behaviorism \cite{Skinner1938TheAnalysis} which describes how agents improve over time by taking actions in an environment with a goal of maximizing \textit{expected return}, the cumulative future reward signal received by the agent \cite{Sutton1998ReinforcementIntroduction}. The control policy of the agent is iteratively improved by selecting actions which maximize \textit{return}. RL problems are often described as sequential decision making problems modelled as Markov Decision Processes (MDPs) which define tuples: $(State, Action, Transitions, \gamma, Reward)$, full details of MDPs are omitted for space and can be found in \cite{Sutton1998ReinforcementIntroduction}. The ultimate goal of an RL agent is to determine a policy which maps a given current state to the correct actions to maximize \textit{expected return}. In this work we use a continuous actor-critic (AC) algorithm (Algorithm \ref{alg:cacrl}) similar to that described in \cite{Pilarski2011OnlineLearning,Pilarski2013Real-timeJoints,Mathewson2016SimultaneousLearning, DBLP:journals/corr/MathewsonP17}. AC methods work to reduce variance in gradient estimation through the use of two learning systems: a policy-focused actor (selects the best action) and a critic (estimate of value function, criticizes actor) \cite{Sutton1998ReinforcementIntroduction}.

The \textit{Interactive Shaping Problem (ISP)} defines the problem of optimizing the incorporation of human feedback into a learning agent in a sequential decision making problem \cite{Knox2010CombiningLearning}. The \textit{ISP} asks: how can the agent learn the best possible task policy, as measured by task performance or cumulative human feedback, given the information contained in the human responses \cite{Knox2009InteractivelyFramework,Knox2012ReinforcementTasks}. While there are many ways to incorporate human knowledge into a learning system both before and during learning \cite{Thomaz2008TeachableLearners,Chernova2014RobotTeachers,argall2009survey, Knox2015FramingPerformance, fails2003interactive}, this paper focuses on incorporating human feedback directly alongside MDP derived reward in an RL problem.

This work builds on the work of Vien and Ertel, who showed that the human feedback model can be generalized to address the problems associated with periods of noisy and inconsistent human feedback (\citeyear{Vien2013LearningSpaces}). Recent advancements in modelling human feedback with a Bayesian approach have improved on these foundations in discrete environments \cite{Loftin2016LearningLearning}. Most recently, work by \citeauthor{Macglashan2016ConvergentHumans} show that human feedback may be better modelled as an advantage function to handle changes in a feedback strategy over time \cite{Macglashan2016ConvergentHumans} and that these techniques may be extended to the policy space \cite{DBLP:journals/corr/MacGlashanHLPRT17}.

The experimental configuration in this study is similar to that described in similar related work \cite{Pilarski2011OnlineLearning, Mathewson2016SimultaneousLearning}. This study provides novel results for $N=13$ different human trainers, and robust comparison with simulated baselines. We explore the implications of varying the conditions of the user providing control signals, feedback signals, and simultaneous control and feedback signals. To characterize the many latent variables in human feedback and control we also compare these results with simulated human trials. Specifically, we investigate the effects of providing control on feedback, and the effects of providing feedback on control.

\subsection{General Interactive Learning Framework}

Fig. \ref{fig:fig8} details a General Interactive Learning Framework. The system is composed of the classic RL learning diagram \cite{Sutton1998ReinforcementIntroduction}. This is seen in the dotted red box surrounding the agent, environment, as well as the lines for action $a$, state $s$, and reward $r$. Both the agent and the environment have interactive components which the human may affect directly through $a_h$. The human may take actions which affect the dense feedback channel $\bm{F}$ or the dense state signal $\bm{S}$. The human perceives the dense display signal $\bm{D}$, this display may take many different forms including but not limited to visual perception of a screen.

The Feedback Interface serves as a filter through which the humans feedback actions are translated to a real-valued human-based signal $h$, and similarly the State Interface translates human state signals into state information to the agent, $s_h$. Finally, the agent delivers meta-information about the action that is taking, which may not be visible by the environment, in $a^+$, and the environment may be observed directly by the human through meta-state information in $s^+$. The framework offers a formalization for experimentation in interactive RL problem domains.

\section{Methods}
\subsection{Experimental configuration: Aldebaran Nao and Myo Electromyography (EMG) Data}
The experimental set up is shown in Fig. \ref{fig:fig7}. If this configuration is compared to Fig. \ref{fig:fig8}, then it is straightforward how the component pieces fit together. It is composed of a human subject interacting with an agent, in this case the Aldebaran Nao robotic simulation platform (Aldebaran Robotics). The human provides control signals through a state interface, in this case a wireless 8-channel Myo electromyography (EMG) armband (Thalmic Labs), and delivers feedback on the feedback interface (keyboard) which is connected to the learning system, here a MacBook Air (Apple, 2.2 GHz Intel Core i7, 8GB RAM). The system provides the user a view of the simulated robot and logs data for future analysis. All experiments in this paper are performed using a simulated Nao platform. The performance of this experimental set-up to be comparable, albeit with less environmental noise, to simulation and real-world experiments \cite{Mathewson2016SimultaneousLearning}. 

For this study, we investigate three experimental human conditions: 1) providing control signals, 2) providing feedback signals and 3) providing simultaneous control and feedback signals. We include simulated control and feedback runs for comparison. The order of the experimental conditions was randomized by participant to control for any potential training effects on both the EMG control signal and/or reward and punishment feedback signals. By comparing our experimental conditions with two simulated conditions: simulating human feedback and control and simulating strictly control signals, we are able to characterize and vary important latent variables hidden from the agent which impact the learning of the system. 

We report the results of asymptotic, convergent task performance, EMG control signal error when compared to ideal EMG control signal, reward timing (when in the robotic trajectory the user is providing feedback), and total feedback counts (both rewards and punishments). These outcome measures allow for comparison between experimental conditions, and will help to model and simulate the human in future work.


\subsection{Actor-Critic Reinforcement Learning}

In this work we use a continuous actor-critic (AC) algorithm (Algorithm \ref{alg:cacrl}) similar to that described in prior work \cite{Pilarski2011OnlineLearning,Pilarski2013Real-timeJoints,Mathewson2016SimultaneousLearning,DBLP:journals/corr/MathewsonP17}. AC methods can reduce variance in gradient estimation through the use of two learning systems: a policy-focused actor (selects the best action) and a critic (estimate of value function, criticizes actor) \cite{Sutton1998ReinforcementIntroduction}. 

Actor-critic algorithms are a subset of policy gradient based algorithms. In these algorithms, the control policy \(\pi(a|s)\) is a function which defines the probability with which the system will select an action \(a\) in a state \(s\). \(\pi\) is characterized by a vector of parameters \(\mathbf{w} \epsilon \mathbb{R}^n \)which we assume is differentiable in \(\mathbf{w}\) for any state-action pair. The goal of policy-based methods is to find a policy \(\pi\) which maximizes \textit{expected return}. Policy-based methods update the policy parameter vector \(\mathbf{w}\) in the direction of the gradient of the return with respect to \(\mathbf{w}\) \cite{williams1992simple}. This gradient can be estimated from samples of interaction between the learning agent and the environment, \((s_t,a_t,s_{t+1},r_t)\) \cite{NIPS1999_1713}. This gradient estimation technique can have high variance, and thus may require a large number of samples to converge.

Actor-critic (AC) methods aim to reduce this problematic variance by using two learning systems: an \textit{actor} and a \textit{critic}. The actor shapes the policy \(\pi\) and selects the actions, and the critic predicts the \textit{expected return} while following the policy \(\pi\). The critic represents the value function of the current policy \(\pi\) in the state \(s\). While this value function is not known, it can be estimated using temporal difference learning; this estimate is then used to compute the change in the parameter vector \(\mathbf{w}\). Because \(S\) is continuous in the variables defining it, a standard function approximation technique, (i.e. tile coding), is often used to transform the state s into a high-dimensional binary feature vector \(x(s)\). This discretizes the space, and allows for generalization in the learned policies \cite{Sutton1998ReinforcementIntroduction}. Details in this paper are limited to those relevant to the current study and can be found elsewhere \cite{Sutton1998ReinforcementIntroduction, peters2008natural}.

\begin{algorithm}[t]
   \caption{Continuous-Action ACRL}
   \label{alg:cacrl}
\begin{algorithmic}
   \STATE \textbf{initialize:} $\mathbf{w}_\mu, \mathbf{w}_\sigma, \mathbf{v}, \mathbf{e}_\mu, \mathbf{e}_\sigma, \mathbf{e}_\mathbf{v}, s$
  \REPEAT
  \STATE $\mu \leftarrow \mathbf{w}_\mu^T \mathbf{x}(s)$
  \STATE $\sigma \leftarrow \exp[\mathbf{w}_\sigma^T \mathbf{x}(s)]$
  \STATE $a \leftarrow \mathcal{N}(\mu,\sigma^2)$
  \STATE \textbf{take action} $a$, \textbf{observe} $r, s'$
  \STATE $\delta \leftarrow r + \gamma\mathbf{v}^T \mathbf{x}(s') - \mathbf{v}^T \mathbf{x}(s)$
  \STATE $\mathbf{e}_\mathbf{v} \leftarrow \min[1,\lambda_\mathbf{v} \gamma \mathbf{e}_\mathbf{v} + \mathbf{x}(s)]$
  \STATE $\mathbf{v} \leftarrow \mathbf{v} + \alpha_\mathbf{v} \delta \mathbf{e}_\mathbf{v}$
  \STATE $\mathbf{e}_\mathbf{\mu} \leftarrow \lambda_\mathbf{w} \mathbf{e}_\mu + (a - \mu) \mathbf{x}(s)$
  \STATE $\mathbf{w}_\mu \leftarrow \mathbf{w}_\mu + \alpha_\mu \delta \mathbf{e}_\mu$
  \STATE $\mathbf{e}_\mathbf{\sigma} \leftarrow \lambda_\mathbf{w} \mathbf{e}_\sigma + [(a - \mu)^2 - \sigma^2] \mathbf{x}(s)$
  \STATE $\mathbf{w}_\sigma \leftarrow \mathbf{w}_\sigma + \alpha_\sigma \delta \mathbf{e}_\sigma$
  \STATE $s \leftarrow s'$
  \UNTIL {termination criteria is met}
\end{algorithmic}
\end{algorithm}

\section{Experiments}
We present the first user study (N=13) of human-robotic interaction where the human simultaneously delivered both control and feedback signals. Our experimental configuration is similar to that of \citet{Mathewson2016SimultaneousLearning, DBLP:journals/corr/MathewsonP17}. We aim to compare the conditions of a human providing control signals, feedback signals, and both control and feedback signals. By exploring the effects of actual human trainers providing simultaneous control and feedback signals to the RL system during the performance of a self-mirrored movement control task we can properly elucidate differences between the conditions.

The task was designed to require the human to multi-task, but not be overly cognitively demanding. First, the right arm of the Nao is preprogrammed to move in a periodic pattern of flexion and extension at the elbow joint between two stable set points within the allowable joint range. The RL agent controls the left elbow joint by selecting, on each time step, an action. Actions are defined by an angular displacement from the current angle. The agent is attempting to move the left elbow joint so that it matches, on every time step, the movement of the preprogrammed right elbow joint. With this configuration and task we are able to define an optimal policy, which would track the preprogrammed joint movement exactly. From this theoretically optimal trajectory we are able to derive environmental (or MDP) based rewards given a set acceptable angular error threshold. When the RL-controlled left elbow joint is within the angular deviation threshold of the preprogrammed right elbow joint then a reward of $1$ is delivered by the MDP. If the joints differ by more than the angular threshold then a negative relative punishment is delivered. The magnitude of this negative reward is proportional to the absolute difference between the preprogrammed right arm (optimal) and the learner controlled left arm.


The human wore the Myo EMG band on their right forearm to deliver EMG control signals. When the right arm joint of the simulated robot was in flexion the human was instructed to flex their wrist, thereby activating muscles in the forearm. Optimal flexion caused the EMG state signal to tend towards $1$. Conversely, when the robot extended the right elbow joint, the human was instructed to relax their arm. Optimal relaxation caused the EMG state signal to tend towards $0$. This movement pattern created an EMG control signal with large signal to noise ratio and two distinct set control points, human wrist flexion and wrist relaxation.

Similar to the state representation used by \citet{Mathewson2016SimultaneousLearning}, the continuous state space is defined by the filtered, time-averaged, and dimensionally reduced EMG signal, and the angle of the actuated left elbow joint, and is represented with approximation using tile coding \cite{Mathewson2016SimultaneousLearning, DBLP:journals/corr/MathewsonP17}. Parameters were set as follows: $\alpha_v = 0.1/m$, $\alpha_\mu = \alpha_\sigma$, $\gamma = 0.9$, $\lambda_w = 0.3$, $\lambda_v = 0.7$, joint angles were limited by manufacturer specifications at $\theta \in [0.0349, 1.5446]$ rads. Weight vectors $\mathbf{w}_\mu, \mathbf{w}_\sigma, \mathbf{v}, \mathbf{e}_\mu, \mathbf{e}_\sigma, \mathbf{e}_\mathbf{v}$ were initialized to $\mathbf{0}$ and standard deviation was bounded by $\sigma \geq 0.01$. The eligibility trace update for the critic was scaled by $\gamma$ to speed up convergence. Maximum number of time steps = 10400, which provided 13 identical 800 time step waveforms trajectories between the two set points. Learning updates and action selection occurred at ~33 Hz or every ~30 ms. The angular joint deviation threshold was set to $\Delta\theta_{max} = 0.05$ and absolute angular joint updates were clipped to $0.1$ to reduce large angular movements. Actions were selected and performed on every time step.

The ACRL system was trained online with human control and feedback. Control was defined by the EMG signals attached to the right arm of the human. Feedback was delivered as reward $(+1)$ or punishment $(-0.5)$ through the pushing of a button on the laptop keyboard. The design of this unbalanced reward scheme $[-0.5, +1]$ is important as the range of the feedback matches the range of the environment derived punishment $(-1.5, 0.05)$. This provides an effective balance between relative effects of human and environment delivered feedback in the algebraic sum of rewards, $r = r_{human} + r_{MDP}$. On all time steps MDP reward and human reward were directly summed and applied to the learning agent update (Algorithm \ref{alg:cacrl}).

We are interested in the effects of the user providing control and feedback signals both independently, and then simultaneously to the system. As the human is unable to give feedback at every step that an agent takes, we need to account for the fact that after the exact time step a feedback is given there are likely suboptimal states which support the optimal trajectory. With a decay parameter we are able to \textit{smear} the human feedback forward in time. In these experiments $\mathit{smear = 0.01}$ \cite{DBLP:journals/corr/MathewsonP17}. It has been shown that the limited human feedback can be applied across near-optimal state-action pairs, and support the agent learning an optimal solution \cite{Pilarski2011OnlineLearning, knox2012humans}. 

Human experimental conditions were: 1) human EMG control signals with no feedback, 2) human EMG control and feedback signals, 3) human feedback signals with simulated EMG control signals, as well simulated conditions were: 4) simulated feedback signals with simulated control signals, and 5) simulated control signals with no feedback signals. The probability of simulated feedback being given was $0.1$, and the probability of that feedback being correct was $0.8$ \cite{DBLP:journals/corr/MathewsonP17}. Task performance was measured by taking the average mean absolute angular error from the last 5k steps, providing an estimate of asymptotic convergence performance. This was done to compare the experimental results after initial learning and helped to reduce noise intrinsic in early learning. We report several additional experimental results: EMG error over task performance measured as the absolute difference from an ideal EMG control signal, reward timing over the robotic movement trajectory, and total feedback count. Subjects (N=13) gave informed consent to participate and the trial was approved by the human research ethics board.

\section{Results}

\begin{figure}[!ht]
	\centering
    \includegraphics[width=3.3in]{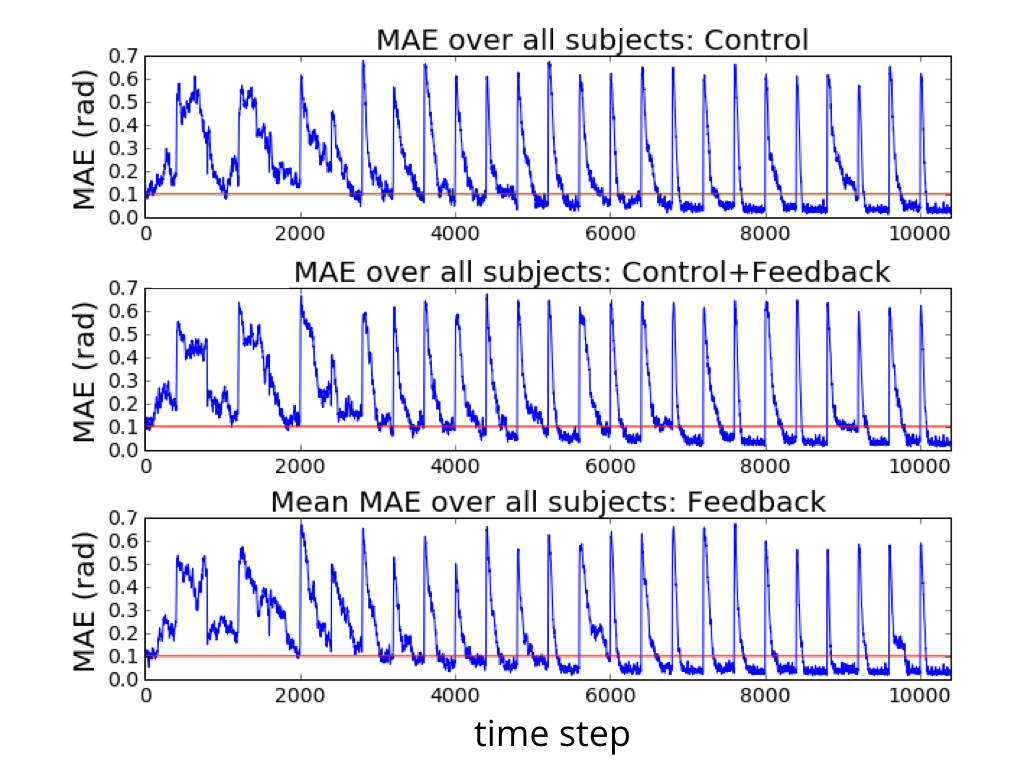}
  	\caption{Performance over Time. MAE (mean angular error) over all subjects (blue), and the ‘target’ threshold (red). This figure shows similar performance across all three experimental conditions (control, control+feedback, feedback), which all manage to find a convergence.}
    \label{fig:fig1}
\end{figure}

This paper presents results of N=13 subjects performing all three human conditions, and N=30 for each simulated condition. Figure \ref{fig:fig1} shows the mean angular error measured over all the subjects and the target threshold is shown in red. This plot indicates that the learning converges for all experimental human conditions. There is no significant difference between the number of time steps before asymptotic convergence between the three conditions. On more difficult learning tasks, such as when the MDP provides constant (instead of proportional) negative rewards, shaping feedback significantly improves learning \cite{Mathewson2016SimultaneousLearning}. The easier learning task in these experiments meant that the system could converge in all runs. If any of the subjects acted purely adversarially, these systems might significantly diverge.

\begin{figure}[t!]
	\centering
    \includegraphics[width=3.3in]{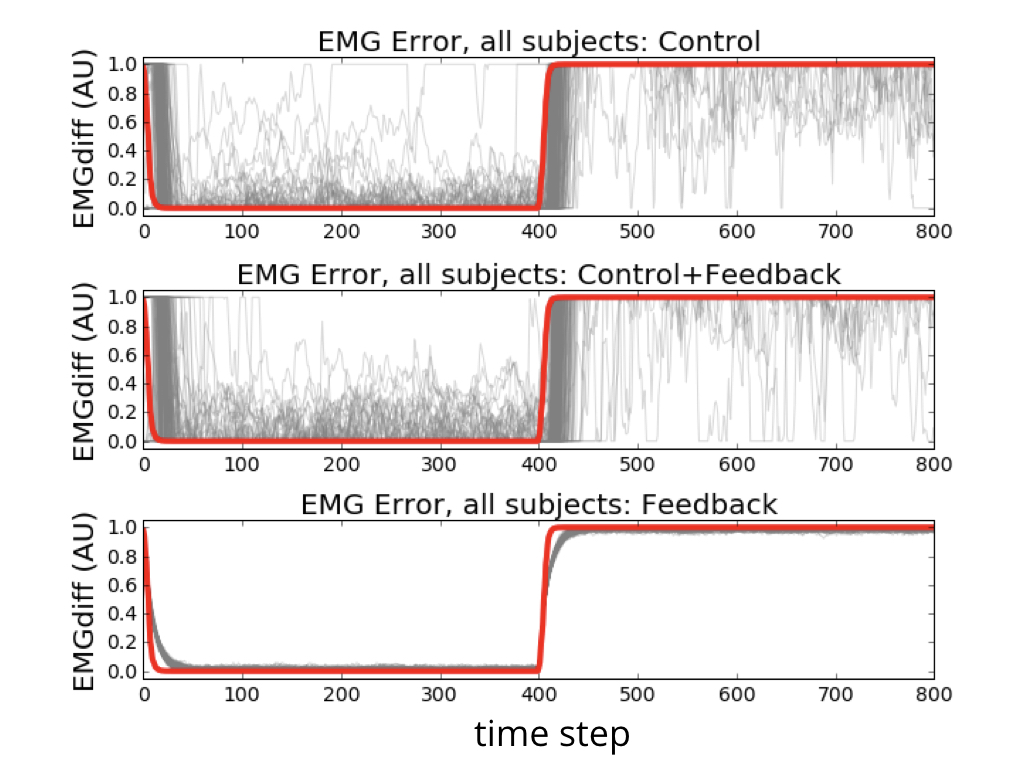}
  	\caption{EMG Errors over Time. EMG error (gray) is measured as the difference between the human-delivered EMG signal and the ideal EMG control signal (red) over all subjects. This figure shows a single window of the control task showing the two set points, low and high. Subject data is reshaped to compare with the ideal EMG control signal in a single window. AU: arbitrary units.}
    \label{fig:fig2}
\end{figure}

Figure \ref{fig:fig2} suggests that there is variability in the EMG control signals delivered by the humans over the trajectory. It suggests that there may be a noticeable increase in the variability of the control signals when the user must simultaneously deliver feedback. The top (Control) and middle (Control+Feedback) figures show that the human control signal is often noisy with artifacts which deviated from the ideal EMG signal. The bottom (Feedback) plot shows the simulated EMG (used in the human feedback condition); by design this signal is relatively stable, with low variability and a comparative lack of outliers. Importantly, the top two plot have thick vertical gray sections around time steps 5-25 and 400-420; this indicates that there is a short reaction window between robotic control and human response. This reaction delay is about 20 time steps, which corresponds to ~660ms, which is well within the human reaction time window \cite{AAAIsymp09-knox, hockley1984analysis}.

\begin{figure}[t!]
	\centering
    \includegraphics[width=3.3in]{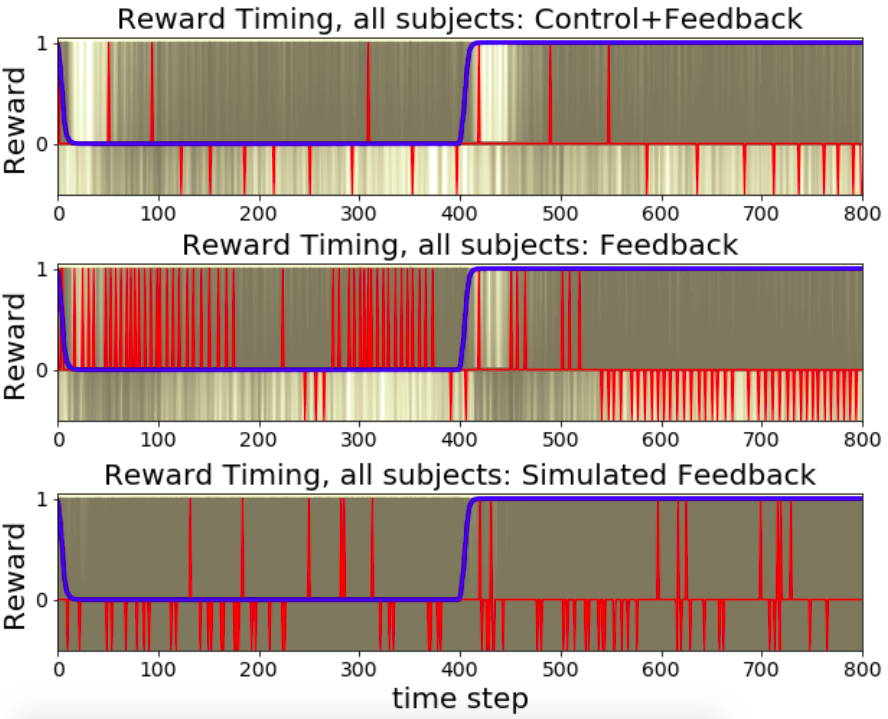}
  	\caption{Reward Timing. Reward timing is plotting all the reward (gray) the human subjects delivered to the robotic control system over the entire run. The red line is a single example from a single subject over a single window. The blue line is the target joint angle, illustrating the movement trajectory.}
    \label{fig:fig3}
\end{figure}

Figure \ref{fig:fig3} plots the timing of the delivered feedback, both reward and punishment. In the top (Control+Feedback) and middle (Feedback) plots, human reward delivered is plotted in gray. When the target joint moves (time step 400-450, visible in blue), the user is temporarily distracted, focuses on the joint movement, and corrects their hand position to ensure that the EMG control signal quickly reflects the target joint angle. During this time, there is a noticeable drop in the amount of feedback that the user is giving to the system (in gray). A single run, over a single movement trajectory window, is plotted in red. This randomly selected example is characteristic of the population effects. The difference between feedback delivered in the top and middle subplots indicates that the feedback is more sparse during simultaneous control and feedback. In the bottom (Simulated Feedback and Control) plot, simulated feedback is delivered with a set probability and thus it is much more consistent across the trajectory, and there is less of a distraction effect.


\begin{figure}[t!]
	\centering
    \includegraphics[width=3.3in]{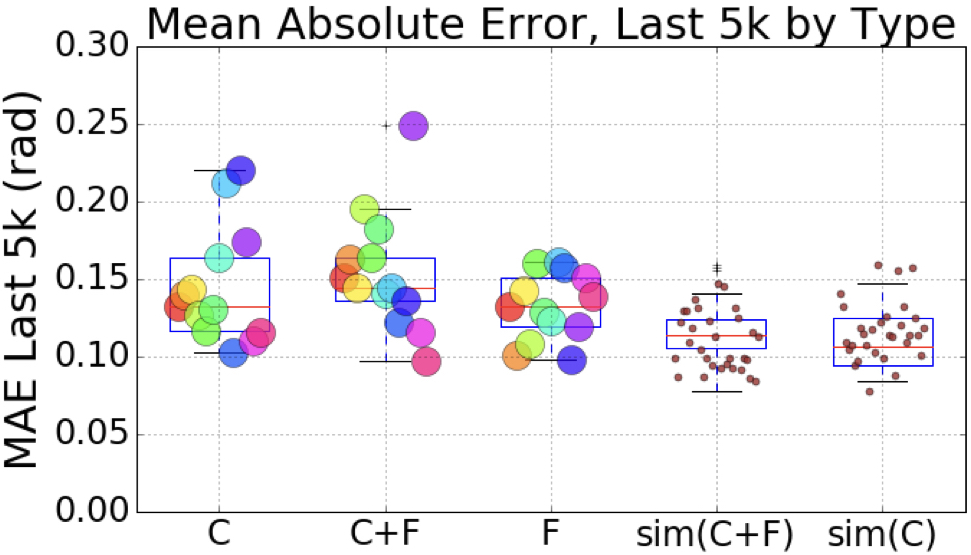}
  	\caption{Mean Absolute Error Variability by Experimental Condition. Mean absolute error (MAE) over the last 5000 time steps by experimental condition. This barplot shows the median and spread of the MAE over the experimental conditions: C:control, C+F:control and feedback, F:feedback, sim(C+F):simulated control and feedback, sim(C):simulated control.}
    \label{fig:fig4}
\end{figure}

Figure \ref{fig:fig4} shows the variability in mean absolute error over the last 5000 time steps by experimental condition. For the first three (human) conditions (Control, Control+Feedback, Feedback), the mean value for each subject is plotted in their own color (consistently offset from left to right), so that subject trends can be traced between conditions. For the last two (simulated) conditions, the mean value of each run is plotted in brown so that the spread around the box plot can be fully appreciated. This plot suggests that given the experimental set up in this paper, there is no significant difference between the human experimental conditions in terms of MAE over the last 5000 time steps. Interestingly, by tracing individual subjects, it can be noted that the performance of some individuals decreases when the subject must perform simultaneous feedback and control. This is qualitatively supported by the anecdotal subject observations of several of the subjects that \textit{concentrating on both control and feedback is difficult}. Additional comments from human participants are included in Supplementary Material. Earlier work would indicate that on harder learning tasks, such as when the negative punishment is constant and not proportional to the distance between the learner joint angle and the optimal joint angle, human feedback can significantly improve performance \cite{Mathewson2016SimultaneousLearning}. It is important to note that the task performance is somewhat limited by the setting of the angular difference threshold, $\Delta\theta_{max}$. The performance of the simulations indicates that this threshold and the maximum number of time steps were limiting factors in asymptotic convergence.

\begin{figure}[t!]
	\centering
    \includegraphics[width=3.3in]{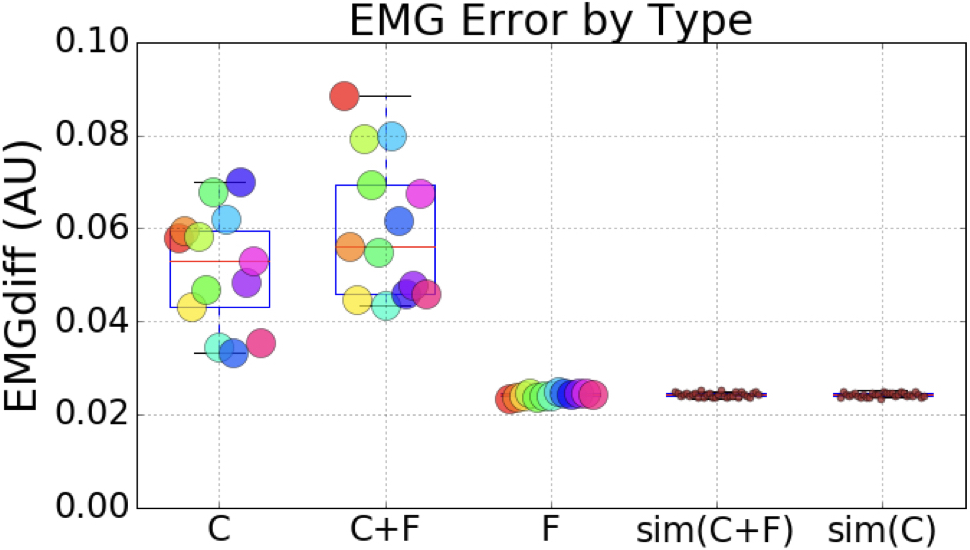}
  	\caption{EMG Error by Experimental Condition. This barplot shows the median and spread of the EMG Error over the human experimental conditions: C:control, C+F:control and feedback, F:feedback, and the simulated experimental conditions: sim(C+F):simulated control and feedback, sim(C):simulated control. Colors correspond exactly to Fig. \ref{fig:fig4}. AU: arbitrary units.}
    \label{fig:fig5}
\end{figure}

Figure \ref{fig:fig5} plots the EMG error by experimental condition. In this task there is no significant difference between the human experimental conditions in terms of EMG error, but there is a subject-specific trend which suggests that there is more variability when the human is providing control and feedback. If the human simultaneously provides feedback and control signals, they take longer to modify their EMG control signals and may make mistakes in the delivery of control signals. It is important to note that because conditions F, sim(C+F), sim(C) rely on simulated EMG data they have consistent, minimal variability in the EMG error. The error that is present is based on the error in the simulated EMG signal which can be visualized in the bottom plot of Fig. \ref{fig:fig2} as the difference between the red (ideal) and gray (simulated) control signals.

\begin{figure}[t!]
	\centering
    \includegraphics[width=3.3in]{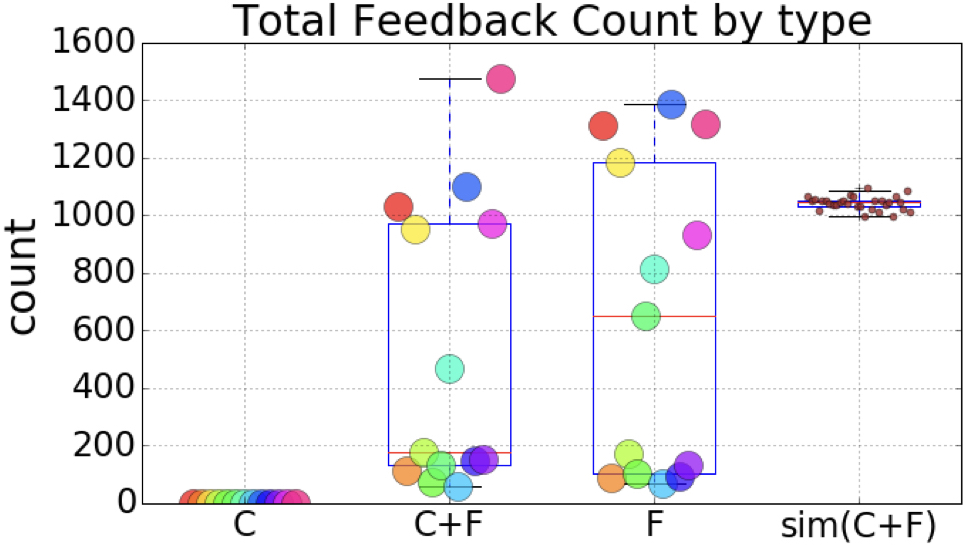}
  	\caption{Total Feedback Count by Experimental Condition. This barplot shows the median and spread of the total feedback given over human experimental conditions: C:control, C+F:control and feedback, F:feedback, and the simulated experimental conditions: sim(C+F):simulated control and feedback.}
    \label{fig:fig6}
    \vspace{-0.5em}
\end{figure}

Figure \ref{fig:fig6} shows the total feedback count by experimental condition. In this task there is no significant difference between the distributions of human experimental conditions in terms of total feedback count, but there is a subject-specific trend which suggests that the human will provide more feedback when they do not have to simultaneously provide control signals. Furthermore, this figure suggests that there may be at least two different styles of users: those that provide plenty of feedback (more than 400), and those who only provide a limited amount of feedback (less than 400). This is an important human-specific teaching style tendency that will require future work to explore. More feedback can provide additional shaping toward optimal goal states thereby improving performance, but has a higher probability of creating positive reward cycles \cite{Ng:1999:PIU:645528.657613}. By tracing individuals by color on Fig. \ref{fig:fig6}, it is evident that human subjects provided the most feedback also performed more optimally on the learning task, on Fig. \ref{fig:fig4}. It is important to note that because condition C incorporates no human feedback, values for all the subjects is 0. Because the simulated condition uses a fixed probability of delivering feedback, in this case $P(feedback)=0.1$ there is not much variability in the simulation runs, resulting in a tight grouping around the boxplot for Simulated Control+Feedback (right most boxplot). It is clearly evident that each run sees a total feedback count of about 1040, or 10\% of the total time steps in each run.

\section{Discussion}

This paper presents the first human user study exploring combinations of human and environmental feedback in a robotic systems with human-derived EMG control signals. Simulation and physical experimentation have shown agreement in past studies. Some performance decrease may be expected with physical robotic systems due to heat, inertia, and mechanical jitter. Simulation provides safety, controlled repeatability and batch processing \cite{Mathewson2016SimultaneousLearning}.


The question of optimal outcome metrics for measuring humans-machines interaction remains open. In this study we explore the effects of tightly coupled control signals on feedback, and the effects of rapidly delivered feedback signals on control signals. In an optimal collaboration, the human should not be required to deliver signals (control and feedback) when the system has no use for them. Similarly, the system should start to model the human delivered signals \cite{Knox2009InteractivelyFramework,Vien2013LearningSpaces}.

There exist a limited number of channels through which a human may provide control and feedback. The field will progress to explore the density of information for these communication channels, as described in General Interactive Learning Framework in Fig. \ref{fig:fig8}. Bandwidth and explicitness in human signals delivered to learning agents has been explored previously \cite{FSS125496,pilarski2015prosthetic}. Future work will quantify amounts of human input input relative to levels of autonomy \cite{smith2013summary}. 



There are noticeable teaching style differences between at least 2 groups of human participants, see Fig. \ref{fig:fig6}, as has been previously explored in related work \cite{loftin2016learning,Macglashan2016ConvergentHumans}. Trainers may also vary on whether they provide more positive or more negative feedback and whether they believe the feedback to be instructional and guiding or responsive and reactive \cite{NIPS2016_6413}. Future algorithmic improvements should exploit predicted teacher style to build beliefs about the provided feedback \cite{loftin2016learning, Macglashan2016ConvergentHumans}. 




By imposing a feedback budget on the human, different teaching styles may be elucidated \cite{torrey2013teaching}. The paradigm of teaching-on-a-budget may also help to reduce feedback fatigue. The amount of feedback, the rate it is given, and when in a training paradigm is delivered may be significantly affected if there is a budget or cost (real or assumed) for delivered feedback. Feedback characteristics may change over training (e.g. amount of feedback given, reward or punishment, correct or incorrect) due to training effects or fatigue. Algorithms should be developed which are robust to, and take advantage of, these deviations.

Future work will explore the correlation between feedback delivered and the absolute error between the controlled joint and the target joint. This correlation may vary over time as the human becomes more knowledgeable about the learning task and more fatigued. Our results indicate that humans apply feedback at moments of steady state, and wait for the robot to find stability before feedback is delivered. From qualitative subject feedback, it seems as though the humans are 'training for stability first', and then 'focusing on the changes between set points'. By investigating a more fluid movement task, with multiple set points, or a completely fluid movement pattern with no set points, the effects on feedback delivery may be more pronounced.

\section{Conclusions}


The main contributions of this paper are two-fold. First, we present the first human user study on a human-robot cooperative task where the human is simultaneously providing control and feedback signals in the training of an actor-critic RL robotic agent. Second, we present a General Interactive Learning Framework (Fig. \ref{fig:fig8}), useful for clearly framing and understanding communication channels in IML systems. 

Our results indicate that feedback changes when control signals are simultaneously delivered, and feedback can be delivered without sacrificing quality of control signals. As well, we show results on how the delivered feedback changes over time during training.

This work provides novel results and a new viewpoint on the human training of a semi-autonomous robotic system, and therefore takes important steps toward IML wherein collaborative human and machine intelligence will be able to solve complex problems of the future. Future work will explore extensions from this experimental platform, methods to model, generalize, and deliver consistent human feedback, and work on end-to-end training of these systems through the incorporation of improved representation learning. We expect that key limitations in human-robot interaction can be addressed by incorporating continually adapting control signals and shaping through feedback, especially when robot capacity exceeds control capabilities. 


\bibliography{paper}

\begin{thebibliography}{33}
\providecommand{\natexlab}[1]{#1}
\providecommand{\url}[1]{\texttt{#1}}
\expandafter\ifx\csname urlstyle\endcsname\relax
  \providecommand{\doi}[1]{doi: #1}\else
  \providecommand{\doi}{doi: \begingroup \urlstyle{rm}\Url}\fi

\bibitem[Argall et~al.(2009)Argall, Chernova, Veloso, and
  Browning]{argall2009survey}
Argall, Brenna~D, Chernova, Sonia, Veloso, Manuela, and Browning, Brett.
\newblock A survey of robot learning from demonstration.
\newblock \emph{Robotics and autonomous systems}, 57\penalty0 (5):\penalty0
  469--483, 2009.

\bibitem[Castellini et~al.(2014)Castellini, Artemiadis, Wininger, Ajoudani,
  Alimusaj, Bicchi, Caputo, Craelius, Dosen, Englehart, Farina, Gijsberts,
  Godfrey, Hargrove, Ison, Kuiken, Markovi{\'{c}}, Pilarski, Rupp, and
  Scheme]{Castellini2014ProceedingsElectromyography}
Castellini, Claudio, Artemiadis, Panagiotis, Wininger, Michael, Ajoudani,
  Arash, Alimusaj, Merkur, Bicchi, Antonio, Caputo, Barbara, Craelius, William,
  Dosen, Strahinja, Englehart, Kevin, Farina, Dario, Gijsberts, Arjan, Godfrey,
  Sasha~B., Hargrove, Levi, Ison, Mark, Kuiken, Todd, Markovi{\'{c}}, Marko,
  Pilarski, Patrick~M., Rupp, Rüdiger, and Scheme, Erik.
\newblock {Proceedings of the first workshop on peripheral machine interfaces:
  Going beyond traditional surface electromyography}.
\newblock In \emph{Frontiers in Neurorobotics}, volume~8, pp.\ ~22, 8 2014.
\newblock ISBN 1662-5218 (Electronic) 1662-5218 (Linking).
\newblock \doi{10.3389/fnbot.2014.00022}.

\bibitem[Chernova \& Thomaz(2014)Chernova and
  Thomaz]{Chernova2014RobotTeachers}
Chernova, Sonia and Thomaz, Andrea~L.
\newblock {Robot Learning from Human Teachers}.
\newblock \emph{Synthesis Lectures on Artificial Intelligence and Machine
  Learning}, 8\penalty0 (3):\penalty0 1--121, 4 2014.
\newblock ISSN 1939-4608.
\newblock \doi{10.2200/S00568ED1V01Y201402AIM028}.

\bibitem[Fails \& Olsen~Jr(2003)Fails and Olsen~Jr]{fails2003interactive}
Fails, Jerry~Alan and Olsen~Jr, Dan~R.
\newblock Interactive machine learning.
\newblock In \emph{Proceedings of the 8th international conference on
  Intelligent user interfaces}, pp.\  39--45. ACM, 2003.

\bibitem[Ho et~al.(2016)Ho, Littman, MacGlashan, Cushman, and
  Austerweil]{NIPS2016_6413}
Ho, Mark~K, Littman, Michael, MacGlashan, James, Cushman, Fiery, and
  Austerweil, Joseph~L.
\newblock Showing versus doing: Teaching by demonstration.
\newblock In Lee, D.~D., Sugiyama, M., Luxburg, U.~V., Guyon, I., and Garnett,
  R. (eds.), \emph{Advances in Neural Information Processing Systems 29}, pp.\
  3027--3035. Curran Associates, Inc., 2016.

\bibitem[Hockley(1984)]{hockley1984analysis}
Hockley, William~E.
\newblock Analysis of response time distributions in the study of cognitive
  processes.
\newblock \emph{Journal of Experimental Psychology: Learning, Memory, and
  Cognition}, 10\penalty0 (4):\penalty0 598, 1984.

\bibitem[Knox \& Stone(2009)Knox and Stone]{Knox2009InteractivelyFramework}
Knox, W.~Bradley and Stone, Peter.
\newblock {Interactively Shaping Agents via Human Reinforcement: The TAMER
  Framework}.
\newblock In \emph{Proceedings of the Fifth International Conference on
  Knowledge Capture}, pp.\  9--16, New York, New York, USA, 2009. ACM Press.
\newblock ISBN 9781605586588.
\newblock \doi{10.1145/1597735.1597738}.

\bibitem[Knox \& Stone(2010)Knox and Stone]{Knox2010CombiningLearning}
Knox, W~Bradley and Stone, Peter.
\newblock {Combining Manual Feedback with Subsequent MDP Reward Signals for
  Reinforcement Learning}.
\newblock In \emph{Proceedings of the 9th International Conference on
  Autonomous Agents and Multiagent Systems}, pp.\  5--12, 2010.
\newblock ISBN 978-0-9826571-1-9.

\bibitem[Knox \& Stone(2012)Knox and Stone]{Knox2012ReinforcementTasks}
Knox, W~Bradley and Stone, Peter.
\newblock {Reinforcement learning from human reward: Discounting in episodic
  tasks}.
\newblock In \emph{The 21st IEEE International Symposium on Robot and Human
  Interactive Communication}, pp.\  878--885, 2012.

\bibitem[Knox \& Stone(2015)Knox and Stone]{Knox2015FramingPerformance}
Knox, W.~Bradley and Stone, Peter.
\newblock {Framing reinforcement learning from human reward: Reward positivity,
  temporal discounting, episodicity, and performance}.
\newblock \emph{Artificial Intelligence}, 225:\penalty0 24--50, 2015.
\newblock ISSN 00043702.
\newblock \doi{10.1016/j.artint.2015.03.009}.

\bibitem[Knox et~al.(2009)Knox, Fasel, and Stone]{AAAIsymp09-knox}
Knox, W.~Bradley, Fasel, Ian, and Stone, Peter.
\newblock Design principles for creating human-shapable agents.
\newblock In \emph{AAAI Spring 2009 Symposium on Agents that Learn from Human
  Teachers}, March 2009.

\bibitem[Knox et~al.(2012)Knox, Glass, Love, Maddox, and Stone]{knox2012humans}
Knox, W~Bradley, Glass, Brian~D, Love, Bradley~C, Maddox, W~Todd, and Stone,
  Peter.
\newblock How humans teach agents.
\newblock \emph{International Journal of Social Robotics}, 4\penalty0
  (4):\penalty0 409--421, 2012.

\bibitem[Loftin et~al.(2016{\natexlab{a}})Loftin, Peng, MacGlashan, Littman,
  Taylor, Huang, and Roberts]{Loftin2016LearningLearning}
Loftin, Robert, Peng, Bei, MacGlashan, James, Littman, Michael~L., Taylor,
  Matthew~E., Huang, Jeff, and Roberts, David~L.
\newblock {Learning behaviors via human-delivered discrete feedback: modeling
  implicit feedback strategies to speed up learning}.
\newblock \emph{Autonomous Agents and Multi-Agent Systems}, 30\penalty0
  (1):\penalty0 30--59, 1 2016{\natexlab{a}}.
\newblock ISSN 15737454.
\newblock \doi{10.1007/s10458-015-9283-7}.

\bibitem[Loftin et~al.(2016{\natexlab{b}})Loftin, Peng, MacGlashan, Littman,
  Taylor, Huang, and Roberts]{loftin2016learning}
Loftin, Robert, Peng, Bei, MacGlashan, James, Littman, Michael~L, Taylor,
  Matthew~E, Huang, Jeff, and Roberts, David~L.
\newblock Learning behaviors via human-delivered discrete feedback: modeling
  implicit feedback strategies to speed up learning.
\newblock \emph{Autonomous Agents and Multi-Agent Systems}, 30\penalty0
  (1):\penalty0 30--59, 2016{\natexlab{b}}.

\bibitem[Macglashan et~al.(2016)Macglashan, Littman, Roberts, Loftin, Peng, and
  Taylor]{Macglashan2016ConvergentHumans}
Macglashan, James, Littman, Michael~L, Roberts, David~L, Loftin, Robert, Peng,
  Bei, and Taylor, Matthew~E.
\newblock {Convergent Actor Critic by Humans}.
\newblock In \emph{International Conference on Intelligent Robots and Systems},
  2016.

\bibitem[MacGlashan et~al.(2017)MacGlashan, Ho, Loftin, Peng, Roberts, Taylor,
  and Littman]{DBLP:journals/corr/MacGlashanHLPRT17}
MacGlashan, James, Ho, Mark~K., Loftin, Robert~Tyler, Peng, Bei, Roberts,
  David~L., Taylor, Matthew~E., and Littman, Michael~L.
\newblock Interactive learning from policy-dependent human feedback.
\newblock \emph{CoRR}, abs/1701.06049, 2017.

\bibitem[Mathewson \& Pilarski(2016)Mathewson and
  Pilarski]{Mathewson2016SimultaneousLearning}
Mathewson, Kory~W. and Pilarski, Patrick~M.
\newblock {Simultaneous Control and Human Feedback in the Training of a Robotic
  Agent with Actor-Critic Reinforcement Learning}.
\newblock In \emph{IJCAI International Joint Conference on Artificial
  Intelligence - Interactive Machine Learning Workshop}, 2016.

\bibitem[Mathewson \& Pilarski(2017)Mathewson and
  Pilarski]{DBLP:journals/corr/MathewsonP17}
Mathewson, Kory~Wallace and Pilarski, Patrick~M.
\newblock Reinforcement learning based embodied agents modelling human users
  through interaction and multi-sensory perception.
\newblock \emph{Accepted to the 2017 AAAI Spring Symposium on Interactive
  Multi-Sensory Object Perception for Embodied Agents, March 27-29, 2017,
  Stanford University, USA.}, 2017.

\bibitem[Ng et~al.(1999)Ng, Harada, and Russell]{Ng:1999:PIU:645528.657613}
Ng, Andrew~Y., Harada, Daishi, and Russell, Stuart~J.
\newblock Policy invariance under reward transformations: Theory and
  application to reward shaping.
\newblock In \emph{Proceedings of the 16th International Conference on Machine
  Learning}, ICML '99, pp.\  278--287, San Francisco, CA, USA, 1999. Morgan
  Kaufmann Publishers Inc.
\newblock ISBN 1-55860-612-2.

\bibitem[Peters \& Schaal(2008)Peters and Schaal]{peters2008natural}
Peters, Jan and Schaal, Stefan.
\newblock Natural actor-critic.
\newblock \emph{Neurocomputing}, 71\penalty0 (7):\penalty0 1180--1190, 2008.

\bibitem[Pilarski \& Sutton(2012)Pilarski and Sutton]{FSS125496}
Pilarski, Patrick and Sutton, Richard.
\newblock Between instruction and reward: Human-prompted switching.
\newblock In \emph{AAAI Fall Symposium Series: Robots Learning Interactively
  from Human Teachers}, 2012.

\bibitem[Pilarski et~al.(2011)Pilarski, Dawson, Degris, Fahimi, Carey, and
  Sutton]{Pilarski2011OnlineLearning}
Pilarski, Patrick~M., Dawson, Michael~R., Degris, Thomas, Fahimi, Farbod,
  Carey, Jason~P., and Sutton, Richard~S.
\newblock {Online human training of a myoelectric prosthesis controller via
  actor-critic reinforcement learning}.
\newblock In \emph{IEEE International Conference on Rehabilitation Robotics},
  pp.\  1--7. IEEE, 6 2011.
\newblock ISBN 9781424498628.
\newblock \doi{10.1109/ICORR.2011.5975338}.

\bibitem[Pilarski et~al.(2013)Pilarski, Dick, and
  Sutton]{Pilarski2013Real-timeJoints}
Pilarski, Patrick~M, Dick, Travis~B, and Sutton, Richard~S.
\newblock {Real-time Prediction Learning for the Simultaneous Actuation of
  Multiple Prosthetic Joints}.
\newblock In \emph{IEEE International Conference on Rehabilitation Robotics},
  pp.\  1--8, 2013.

\bibitem[Pilarski et~al.(2015)Pilarski, Sutton, and
  Mathewson]{pilarski2015prosthetic}
Pilarski, Patrick~M, Sutton, Richard~S, and Mathewson, Kory~W.
\newblock Prosthetic devices as goal-seeking agents.
\newblock In \emph{2nd Workshop on Present and Future of Non-Invasive
  Peripheral-Nervous-System Machine Interfaces, Singapore}, 2015.

\bibitem[Skinner(1938)]{Skinner1938TheAnalysis}
Skinner, Burrhus~Frederic.
\newblock \emph{{The Behavior of Organisms: An experimental analysis}}.
\newblock {Appleton-Century}, Oxford, 1938.
\newblock ISBN 1-58390-007-1.
\newblock \doi{10.1037/h0052216}.

\bibitem[Smith(2013)]{smith2013summary}
Smith, BW.
\newblock Summary of levels of driving automation for on-road vehicles.
\newblock \emph{Center for Internet and Society, Stanford Law School}, 2013.

\bibitem[Sutton \& Barto(1998)Sutton and
  Barto]{Sutton1998ReinforcementIntroduction}
Sutton, Richard~S. and Barto, Andrew~G.
\newblock \emph{{Reinforcement learning: An introduction}}.
\newblock MIT Press, Cambridge, 1st edition, 1998.

\bibitem[Sutton et~al.(2000)Sutton, McAllester, Singh, and
  Mansour]{NIPS1999_1713}
Sutton, Richard~S, McAllester, David~A., Singh, Satinder~P., and Mansour,
  Yishay.
\newblock Policy gradient methods for reinforcement learning with function
  approximation.
\newblock In Solla, S.~A., Leen, T.~K., and M\"{u}ller, K. (eds.),
  \emph{Advances in Neural Information Processing Systems 12}, pp.\
  1057--1063. MIT Press, 2000.

\bibitem[Thomaz \& Breazeal(2008)Thomaz and
  Breazeal]{Thomaz2008TeachableLearners}
Thomaz, Andrea~L. and Breazeal, Cynthia.
\newblock {Teachable robots: Understanding human teaching behavior to build
  more effective robot learners}.
\newblock \emph{Artificial Intelligence}, 172\penalty0 (6-7):\penalty0
  716--737, 4 2008.
\newblock ISSN 00043702.
\newblock \doi{10.1016/j.artint.2007.09.009}.

\bibitem[Thomaz et~al.(2005)Thomaz, Hoffman, and Breazeal]{thomaz2005real}
Thomaz, Andrea~Lockerd, Hoffman, Guy, and Breazeal, Cynthia.
\newblock Real-time interactive reinforcement learning for robots.
\newblock In \emph{AAAI 2005 Workshop on Human Comprehensible Machine
  Learning.}, 2005.

\bibitem[Torrey \& Taylor(2013)Torrey and Taylor]{torrey2013teaching}
Torrey, Lisa and Taylor, Matthew.
\newblock Teaching on a budget: Agents advising agents in reinforcement
  learning.
\newblock In \emph{Proceedings of the 2013 International Conference on
  Autonomous Agents and Multi-Agent Systems}, pp.\  1053--1060. International
  Foundation for Autonomous Agents and Multiagent Systems, 2013.

\bibitem[Vien et~al.(2013)Vien, Ertel, and Chung]{Vien2013LearningSpaces}
Vien, Ngo~Anh, Ertel, Wolfgang, and Chung, Tae~Choong.
\newblock {Learning via human feedback in continuous state and action spaces}.
\newblock \emph{Applied Intelligence}, 39\penalty0 (2):\penalty0 267--278, 9
  2013.
\newblock ISSN 0924669X.
\newblock \doi{10.1007/s10489-012-0412-6}.

\bibitem[Williams(1992)]{williams1992simple}
Williams, Ronald~J.
\newblock Simple statistical gradient-following algorithms for connectionist
  reinforcement learning.
\newblock \emph{Machine Learning}, 8\penalty0 (3-4):\penalty0 229--256, 1992.

\end{thebibliography}
\bibliographystyle{icml2017}

\end{document}